\documentclass{article}

\usepackage{arxiv}

\usepackage[utf8]{inputenc} 
\usepackage[T1]{fontenc}    
\usepackage{hyperref}       
\usepackage{url}            
\usepackage{booktabs}       
\usepackage{amsfonts}       
\usepackage{nicefrac}       
\usepackage{microtype}      
\usepackage{graphicx}
\usepackage{amsmath}        
\graphicspath{ {./images/} }

\hypersetup{
    colorlinks=true,
    linkcolor=black,
    citecolor=black,
    urlcolor=black
}

\title{Grounded Inference: Principles for Deterministically Encapsulated Generative Models}

\author{
 Marty ~O'Neill, ~Ph.D. \\
 Odenton, MD, USA \\
  \texttt{martyoneillii@gmail.com} 
}

\begin{document}
\maketitle

\begin{abstract}
The incorporation of generative models into traditional computational systems presents both enormous opportunity and tremendous peril. Although many early adopters have realized these perils at great expense, the field still requires foundational frameworks to de-risk incorporation of AI into traditional systems. This manuscript establishes this foundation through the definition of four specific primitives of AI blended architecture, designed to enable deterministic encapsulation of probabilistic models. It further establishes two overarching anti-patterns broadly represented across industry to serve as warnings for engineers in this field. This framework was designed to enable successful integration of AI into traditional systems while providing a foundation upon which generative model providers could build the next generation of generative model interfaces. 
\end{abstract}

\section{\textbf{Introduction}}

Traditional computational systems rely on strict interface and schema contracts to manufacture a structurally predictable runtime environment out of inherently non-deterministic physical infrastructure. Generative models such as Large Language Models (LLMs) do not conform to this pattern and are \textbf{deceptively chaotic}. Although they provide an illusion of state and control, they are functionally stateless engines that operate via suggestion rather than command. Thus, their internal behaviors cannot be strictly guaranteed. Since generative models cannot guarantee absolute contract compliance, when they are applied within systems that demand this compliance, they must be isolated inside explicit runtime encapsulation frameworks and deterministic guardrails to force their probabilistic outputs into predictable enterprise structures. Although prompt engineering can increase the likelihood of valid results, this practice cannot guarantee adherence to contracts. Nonetheless, the illusion of state and control persist, leading to failures along the following three dimensions:

\begin{itemize}
    \item \textbf{Reproducibility:} There is a pervasive assumption that fixing the random seed and setting the temperature to zero forces a deterministic pipeline. While a temperature of zero forces greedy decoding, the physical execution layer of parallel GPU environments still introduces non-deterministic variance. Because floating-point operations are non-associative, dynamic parallel thread execution alters the precision of matrix aggregations. Achieving exact bitwise reproducibility across enterprise clusters is practically infeasible without sacrificing the parallelization required for inference at scale.
    \item \textbf{Explainability:} Traditional execution paths can be definitively audited through stack traces, conditional logic, and deterministic logs. A probabilistic engine, however, resolves instructions through statistical weights rather than discrete logic trees. When tasked with routine data manipulation, its internal routing cannot be explicitly mapped or mathematically proven, making it a liability unless its deployment is strictly isolated from the core data processing pipeline.
    \item \textbf{Constraint:} Engineers are trained to issue strict, immutable commands to traditional infrastructure. A prompt, however, is not a command; it is an injection of semantic context that shifts the probability distribution of the next token. The engine does not inherently respect strict execution boundaries, and schema constraints or formatting instructions can be statistically overridden if the underlying context window heavily biases an alternative generation path.
\end{itemize}
The central challenge of modern systems architecture is not maximizing the cognitive capabilities of generative models, but securely bounding them. Integrating generative models into production pipelines requires a rigorous definition of the execution boundary: the precise interface where traditional compute ends and generative inference begins. 

A common architectural error is assuming that the input or output data format defines the execution boundary. For example, a traditional script can ingest structured database records and inject them into a parameterized natural language template. While the output is unstructured prose, the execution path is entirely deterministic. The architectural boundary is therefore dictated not by the format of the data, but by the mechanics of the transformation logic. 

Generative models must only be introduced when the transformation requires probabilistic pattern synthesis (e.g., semantic reasoning) that cannot be codified via deterministic algorithms or rigid rules. By isolating probabilistic variance strictly to blocks requiring semantic reasoning, enterprise frameworks can enforce rigid external boundaries. Encapsulating probabilistic engines within deterministic logic ensures that they are applied minimally, only to bridge gaps infeasible through deterministic methods. 

To enable this discussion, we categorize data into two fundamental states:

\begin{itemize}
    \item \textbf{Order (Structured Data):} Information constrained by a rigid, predefined schema. It is deterministic and parseable by traditional computing logic (e.g., typed JSON, SQL tables, ASTs). Crucially, data that appears unstructured at first glance but can be effectively and consistently parsed using deterministic methods (e.g., regular expressions) is, by definition, Order.
    \item \textbf{Entropy (Unstructured Data):} Information lacking a rigid schema, requiring semantic reasoning to derive value (e.g., natural language prose or multimodal audio).
\end{itemize}

In real-world systems, data are frequently blended. A rigid API payload may contain a free-text "notes" field originally meant to be parsed by humans. To handle such cases, engineers must immediately apply the principle of \textbf{Entropy Minimization}. At the earliest possible stage of the ingestion pipeline, engineers must extract all strictly structured elements from the blended payload and route them to traditional compute. The generative model must only be invoked to process the specific fragments requiring semantic reasoning. This strictly minimizes the system's surface of exposure to probabilistic failure modes.

Using the probabilistic engine as a data worker must be avoided when possible. If the source data are comprised entirely of organic entropy such that a deterministic parser cannot be constructed, the engineer may choose to use the probabilistic engine as an ephemeral transformation node. However, the model must not become part of the data plane. Instead, model output must immediately be deterministically validated.

This manuscript does not attempt to catalog every possible design pattern blending generative models with traditional systems. Instead, it defines the underlying physics of generative architecture (The Atomic Primitives) and provides a definitive Reference Architecture to demonstrate how deterministic encapsulation solves the crisis of state-leaking generative models. Furthermore, the selection of specific large language models and the precise formulation of system prompts fall outside the scope of this text. These are fluid implementation variables left to the discretion of the engineer. Our focus remains exclusively on the architectural structures, data flow patterns, and orchestration boundaries required to safely encapsulate generative models within traditional computational systems. 

\section{\textbf{Grounded Inference Primitives}}

Before exploring specific design patterns, we must define the four Grounded Inference Primitives. Every robust enterprise pattern blending generative models with traditional computational systems is simply a deliberate configuration of these four components.

\begin{enumerate}
    \item The Probabilistic Engine: The generative model (e.g., an LLM or multimodal foundation model). It is strictly a stateless reasoning engine. It must never be used as a database, and its outputs must never be trusted as structurally sound without external validation.
    \item The Model Encapsulator: The traditional compute layer responsible for enforcing "Order". It intercepts the output of the Probabilistic Engine and validates it against a rigid, pre-defined schema or semantic contract. If the output fails, the Encapsulator blocks the payload, generates an error trace, and deterministically decides whether to prompt the Probabilistic Engine to try again. The Encapsulator is the sole component that communicates directly with the Probabilistic Engine. It effectively isolates the Probabilistic Engine away from the State Registry and the Deterministic Orchestrator.
    \item The State Registry: The deterministic database, vector store, or document repository. It holds the immutable truth and context of the system. The State Registry must remain isolated from the Probabilistic Engine, connected only through traditional orchestration and validation layers.
    \item The Deterministic Orchestrator: The traditional commodity compute environment (e.g., Python scripts, Kubernetes clusters, message queues). It manages the state flow, handles API routing, and executes the business logic that connects the other three primitives.
\end{enumerate}

\section{\textbf{Generative Anti-Patterns}}

Two anti-patterns are prominently represented as of the writing of this manuscript. Understanding the peril that exists within these anti-patterns is a critical first-step towards building resilient, reliable, and scalable systems that harness the power of generative models.

\subsection{\textbf{Anti-Pattern A: The Generative Model as a Data Worker}}

This anti-pattern occurs when an engineer routes data payloads that could be processed via traditional algorithms through a Probabilistic Engine, or when an engineer routes raw Entropy through a model without immediately forcing the model output through deterministic validation checks. If possible, the model should be used to synthesize a deterministic script to handle the data rather than treating the model as an active worker on the data plane.

\textbf{Systemic Failures}

\begin{itemize}
    \item \textbf{Loss of Auditability and Explainability:} When deterministic data passes through a Probabilistic Engine, the strict causal link between input and output is severed. Under failure conditions, the engineer cannot trace the logic path to explain why a specific record was altered or dropped. Guarantees of reproducibility are elusive due to many factors including non-associative floating point arithmetic executed across distributed systems leading to rounding errors that can alter results. Even if reproducibility could be guaranteed, the causal link remains severed. This could expose the organization to severe financial and legal liability.
    \item \textbf{The Model-as-a-Service Vulnerability:} Modern generative models are typically accessed via third-party APIs. Even if an organization accessing the generative model operates within the same overarching enterprise ecosystem as the organization that provides it, the underlying inference algorithms, routing layers, and model weights can be updated by the provider without notice. Consequently, commonly used methods of enforcing reproducibility (e.g., temperature settings or controlling pseudorandom seeds) are unreliable.
    \item \textbf{Silent Degradation:} The model will inevitably hallucinate or semantically alter data values during transformation. Because the processing occurs at scale within a black box, this corruption enters the downstream database silently, potentially leading to catastrophic results prior to detection.
    \item \textbf{Concurrency Bottlenecks:} Inference within generative models is orders of magnitude slower and more expensive than traditional compute. Forcing billions of records through an API endpoint destroys system throughput and balloons operational costs.
    \item \textbf{Probabilistic Control Flow:} Engineers frequently attempt to utilize generative models to execute conditional routing logic based on the semantic evaluation of incoming payloads. Because prompts are not deterministic conditional statements, the routing path is subject to statistical hallucination, leading to misrouted or orphaned data streams that bypass deterministic exception handlers. 
\end{itemize}

\subsection{\textbf{Anti-Pattern B: The Generative Model as a Stateful Machine}}

This anti-pattern occurs when an engineer relies on a model's context window to act as a rolling memory store or state registry across multiple interactions. Whether the system is attempting to track specific variable values through consecutive prompts, or holding the state of a growing output (such as a lengthy document or complex codebase) for iterative regeneration, the model is erroneously treated as a stateful machine. Generative models lack true internal state management. Therefore, provided values, constraints, or historical contexts can be silently lost or arbitrarily modified without warning as the context window shifts.

\textbf{Systemic Failures}
\begin{itemize}
    \item \textbf{Silent Modification of Frozen Assets:} As the interaction lengthens or the output grows, the model will begin to alter text, data, or variable states that the engineer intended to remain static. This occurs due to semantic drift or context window limitations forcing the model to aggressively summarize or drop earlier sections. An example of such processing degradation has been observed and described by Liu et.al. \cite{Liu2024} as conforming to a U-shaped performance curve, where transformer models effectively access information located at the absolute margins (primacy and recency vectors) but suffer severe token processing and retrieval degradation when critical structural criteria sit in the middle of long input contexts.
    \item \textbf{Context Collapse:} Without external state management, the model eventually reaches a critical mass where it can no longer balance the historical context with the current instruction, leading to catastrophic formatting failures or total task abandonment.
\end{itemize}

To ground the effects of Anti-Pattern B in traditional systems architecture, relying on an LLM context window to maintain state is equivalent to executing code within a runtime environment that arbitrarily and silently alters variable scopes during execution. In a statically typed, compiled language, such as C, a variable declared within a block constraint is guaranteed to retain its allocated bitwise state until it explicitly reaches its end of scope or is overwritten by deterministic logic. Conversely, injecting an instruction such as x=5 into an LLM context provides no allocation guarantees. The token string is not compiled into a stable memory address. It is merely an entry in an attention weight matrix. When the model fails to retrieve these buried operational constraints due to this U-shaped degradation \cite{Liu2024} , it compensates by statistically hallucinating replacement values. Consequently, rather than throwing a null pointer or an out-of-scope exception, the probabilistic engine returns a plausible but entirely fabricated value for x. This mechanism is manifested as silent state mutation to the downstream system, corrupting the execution environment without triggering standard fault handlers.

\section{\textbf{The Encapsulation Micro-Pattern}}
\begin{figure}
    \centering
    \includegraphics[width=0.3\linewidth]{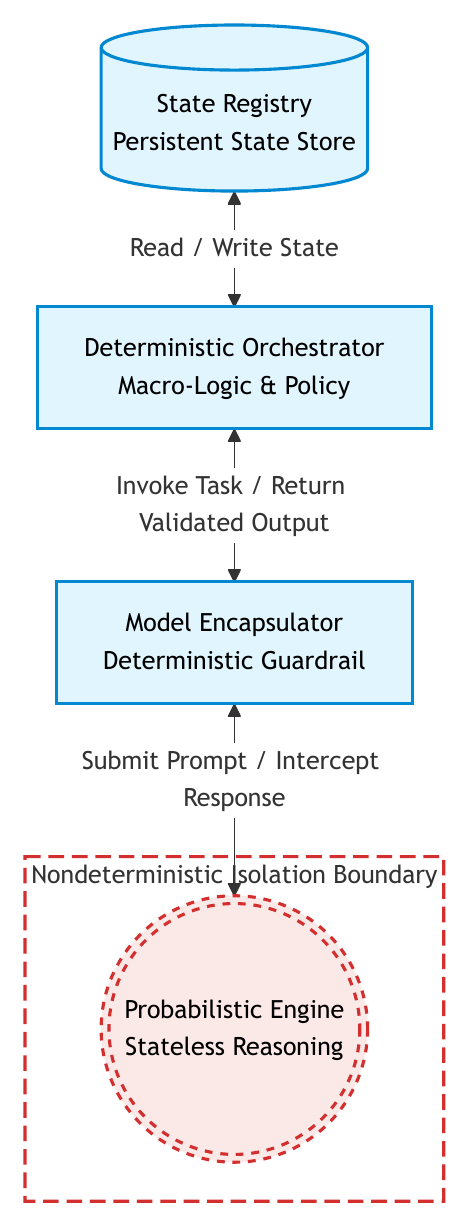}
    \caption{The Encapsulation Micro-Pattern}
    \label{fig:emp}
\end{figure}

This manuscript frames its design patterns in the example of a reference architecture to implement secure data ingestion, transformation, and edge defense. The reference architecture was designed around the Encapsulation Micro-Pattern depicted in Figure~\ref{fig:emp} to provide a structurally robust method for an agent to extract executable logic from a Probabilistic Engine without allowing unpredictable variance to infect the broader distributed system. This architecture explicitly rejects the paradigm of intrinsic self-correction, the assumption that a generative model can reliably identify and correct its own logical errors through raw semantic self-reflection.

This rejection is explicitly grounded in the empirical findings of Huang et.al. \cite{Huang2024}, as well as the Self-Debugging paradigms established by Chen et.al. \cite{Chen2024}. Both frameworks demonstrate that without an objective external anchor, generative models struggle to correct their own reasoning, frequently flipping initial correct assertions into incorrect states. Conversely, the architectural deployment of a compiler or runtime executor running automated test blocks within an isolated sandbox provides a reliable validation ecosystem. By shifting the evaluative burden away from the model’s internal weights and onto an unyielding execution environment managed by the Model Encapsulator, error telemetry is transformed from non-deterministic prose into explicit execution traces.

Because agents implementing this design pattern operate asynchronously within an offline remediation pipeline, compute efficiency and probabilistic error mitigation are the primary constraints rather than real-time latency. To enforce these constraints, the execution boundary is strictly maintained by the Model Encapsulator. This primitive encapsulates the Probabilistic Engine execution, isolating it from the rest of the architecture, maintaining the localized state of the transaction, and enforcing a strict, deterministic retry budget (e.g., a maximum of 3 attempts). The Deterministic Orchestrator remains completely divorced from raw prompt iterations, driving downstream state routing solely based on the definitive, validated success or failure payloads returned by the Encapsulator.

If the Probabilistic Engine exhausts its allocated compute budget without generating a payload that satisfies the Model Encapsulator, the Model Encapsulator terminates the loop and returns a hard execution exception to the Deterministic Orchestrator. Upon receiving this validation failure signal, the Deterministic Orchestrator trips the deterministic system circuit breaker, commits the detailed error telemetry to a State Registry (i.e., an Audit Ledger) for human review, and safely halts the agent's execution thread.

\section{\textbf{The Reference Architecture: The Adaptive Resolution Agent}}

\subsection{\textbf{Intent}}

To act as a unified heuristic remediator capable of resolving schema drift by synthesizing new parsing logic from blended contextual assets, deterministically verifying execution syntax within an isolated sandbox, and semantically validating the output against constraints derived from external documentation. 

\subsection{\textbf{Motivation}}

Data pipelines can fracture for a number of reasons, including undocumented operational shifts (e.g., edge sensor degradation) and documented schema mutations (e.g., upstream vendor API changes). The Adaptive Resolution Agent consolidates both data-driven synthesis and document-grounded verification into a unified, single-instance Control Plane entity. By dynamically querying a Document Store, the agent determines its own epistemological constraints. 

\subsection{\textbf{Applicability}}

Use the Adaptive Resolution Agent when:

\begin{itemize}
    \item Upstream data formats are volatile, and the root cause of drift may be a mix of documented API updates and undocumented system anomalies.
    \item The system requires proactive, top-down logic updates when documentation is available, but still demands streamlined, bottom-up synthesized fixes when upstream providers fail to publish change-logs.
    \item The system mandates that all generative synthesis are deterministically encapsulated and validated before promotion to the primary execution plane.
\end{itemize}

\begin{figure}
    \centering
    \includegraphics[width=1\linewidth]{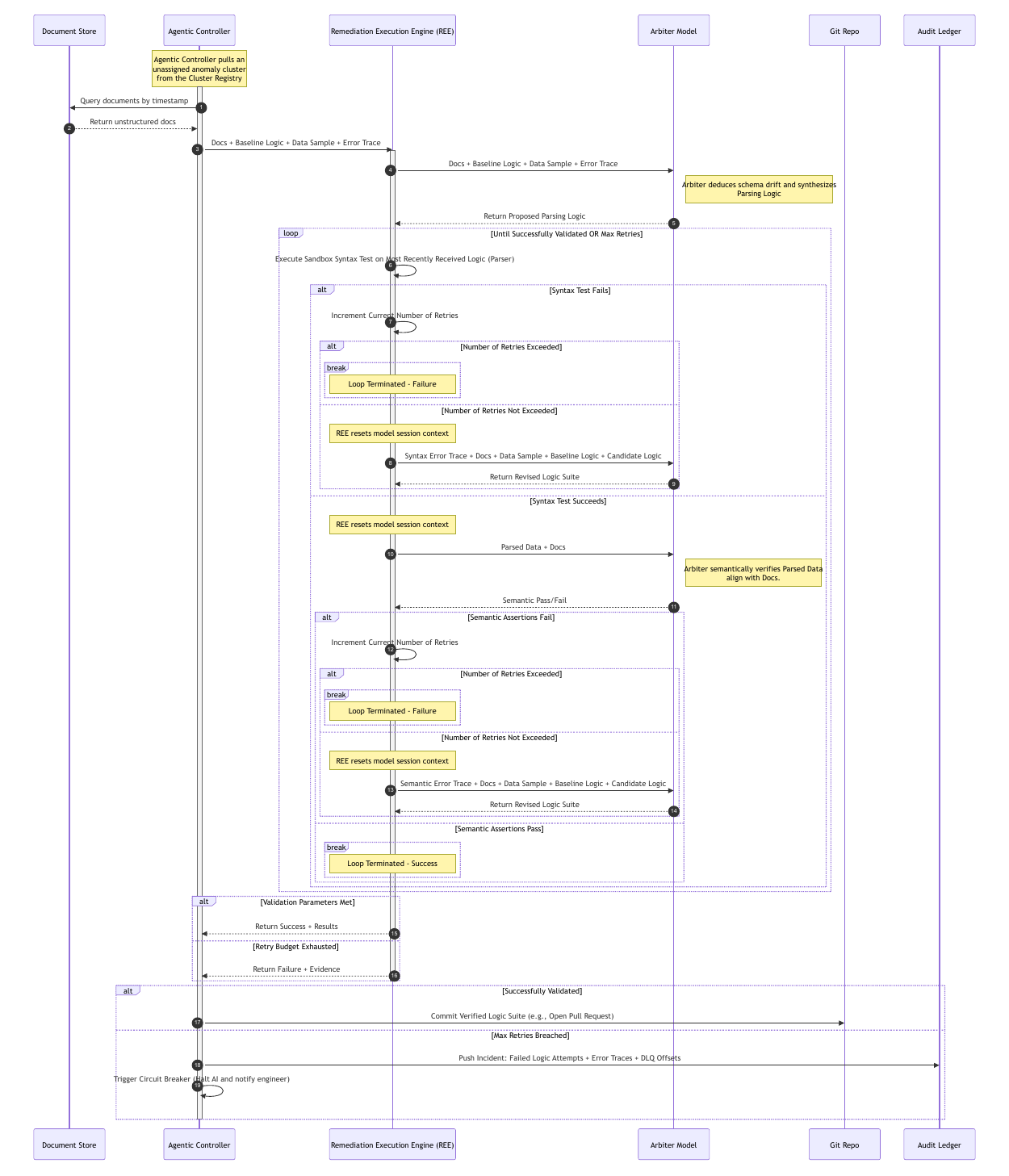}
    \caption{Adaptive Resolution Agent Sequence Diagram}
    \label{fig:ara}
\end{figure}

\subsection{\textbf{Participants}}

\begin{itemize}
    \item \textbf{Document Store (The State Registry):} A domain-specific aggregation of unstructured text. The identification and retrieval of these resources are left to the implementation engineer.
    \item \textbf{Cluster Registry (State Registry):} The database that tracks the lifecycle, groupings, and metadata of anomalies isolated from the primary data plane.
    \item \textbf{Git Repo (State Registry): }The version-controlled destination receives the verified logic suites as commits, entirely decoupling generative synthesis from the standard CI/CD deployment pipeline.
    \item \textbf{Audit Ledger (State Registry): }The repository for capturing failure telemetry and hallucination spirals when the circuit breaker trips. 
    \textbf{Agentic Controller (The Deterministic Orchestrator):} The traditional compute process that interacts with external resources, tasks the Remediation Execution Engine, and writes out-of-band payloads to the Git Repo or Audit Ledger.
    \item \textbf{Remediation Execution Engine (Model Encapsulator):} The traditional compute process that tests the Arbiter Model's proposed logic against a data sample, verifies the parsed data against existing documentation, and controls the looping and conditional logic to enable subsequent retries. It is the only participant that communicates with a Probabilistic Engine.
    \item \textbf{Arbiter Model (Probabilistic Engine):} The unified generative model tasked with deducing schema drift, synthesizing deterministic parsing logic, and semantically verifying data resulting from its synthesized parsing logic.
\end{itemize}

\subsection{\textbf{Collaborations}}

As depicted in Figure~\ref{fig:ara}, the Agentic Controller initiates the remediation process by pulling an unassigned anomaly cluster from the Cluster Registry. It queries the Document Store by timestamp to retrieve the corresponding unstructured documentation or internal wikis associated with the pipeline failure.

The Agentic Controller packages the unstructured documents, the baseline processing logic, the data sample, and the diagnostic error trace, routing the consolidated payload to the Remediation Execution Engine (REE). The REE forwards this context to the Arbiter Model, which analyzes the nature of the schema drift to synthesize a proposed parsing logic suite. The Arbiter Model returns this synthesized logic to the REE.

Upon receipt of the initial candidate logic, the REE then enters a deterministic, bounded retry loop. It first executes a local sandbox syntax test on the most recently received logic suite. If the syntax test fails, the REE increments the current retry count. If the retry budget is breached, the REE breaks the loop. If the retry budget is not exceeded, the REE resets the Arbiter Model session context to maintain statelessness. It then packages the syntax error trace, documentation, data sample, baseline logic, and the failed candidate logic, passing it back to the Arbiter Model for a revised logic suite.

If the syntax test succeeds, the REE transitions to semantic validation. First, it resets the Arbiter Model session context. Next, it routes the resulting parsed data along with the reference documentation back to the Arbiter Model. The Arbiter Model semantically evaluates the parsed data against the documentation constraints and returns a binary semantic pass/fail token to the REE.

If the semantic validation fails, the REE increments the retry count. If retries are exceeded, the REE breaks the loop. If the retry budget remains valid, the REE resets the Arbiter Model session context and submits the semantic error trace, documentation, data sample, baseline logic, and the failed candidate logic back to the Arbiter Model for a revised logic suite. Conversely, if the semantic validation passes, the REE executes an immediate loop break. 

Once the loop terminates, the REE evaluates the final transactional state. If the validation parameters were successfully met, the REE returns a definitive success and results payload to the Agentic Controller. If the retry budget was exhausted, it returns a failure and evidence payload. The REE then deactivates.

If the Agentic Controller receives a successful validation signal, it commits the verified logic suite to the Git repository to open a Pull Request. If the max retry budget was breached, the Agentic Controller pushes the failed logic attempts, error traces, and queue offsets to the Audit Ledger, trips a deterministic system circuit breaker to halt automated threads, and flags the cluster for second level (e.g., by humans) engineering review. 

\subsection{\textbf{Consequences}}

\textit{Benefits:} Provides automated structural adaptation to system drift while streamlining second-level intervention. By anchoring code synthesis in external reference documentation rather than relying on empirical guesswork alone, the architecture enforces verifiable alignment with upstream system realities.

\textit{Liabilities:} Highly susceptible to context window limitations. Processing large volumes of documentation can result in massive token costs or cause the Probabilistic Engine to lose thread of specific logical constraints. As demonstrated by Liu et al. \cite{Liu2024} , increasing the text payload introduces a distinct structural trade-off: providing a model with more out-of-band data simultaneously scales the volume of context it must reason over, predictably lowering information extraction accuracy if targeted documentation mappings settle in the center of the payload.

\subsection{\textbf{Implementation}}

Engineers face a critical context-management challenge. They must execute token count estimation functions across the provided documents to strategically partition the data. Improper splitting breaks the Arbiter Model’s ability to identify cross-document patterns. To mitigate this, the architecture requires token compaction methods to condense documentation payloads before arbitration. Furthermore, documentation sourcing (whether from official standards bodies or organic scraping pipelines) must be explicitly filtered and validated by an authoritative knowledge pipeline to prevent context contamination.

\subsection{\textbf{Related Patterns}}

\textbf{The Generative Model as a Data Worker (Anti-Pattern A):} The fragile design paradigm that the Adaptive Resolution Agent avoids by using the LLM to write the parser, rather than process the data directly.

\section{\textbf{The Resolution of the Anti-Patterns}}

The Reference Architecture presented in this manuscript is not merely a theoretical exercise. It is the direct, mechanical inversion of the generative anti-patterns defined herein. By restricting the Probabilistic Engine strictly to the synthesis of deterministic parsers, the system neutralizes Anti-Pattern A (The Generative Model as a Data Worker). The generative model is physically barred from mutating the primary data stream, isolating stochastic variance to a sandboxed control plane. Further, by delegating all execution state, such as retry budgets, loop management, and context purging, exclusively to the deterministic control layers, the system neutralizes Anti-Pattern B (The Generative Model as a Stateful Machine). The Probabilistic Engine is instantiated as an ephemeral, stateless function that resolves a tightly bounded probabilistic task and immediately terminates. It holds no rolling context, preventing accumulation from causing catastrophic context collapse and state leakage.

The Reference Architecture codifies specific constraints that enable autonomous orchestration features previously unachievable under unstructured generative deployments. If there is a singular takeaway for the engineer, it is that Probabilistic Engines must be treated as strictly stateless machines. A critical design principle is the rejection of the "Information Pass-Through" model (which itself is an instance of Anti-Pattern B: The Generative Model as a Stateful Machine). Only the specific information required for the engine to complete its reasoning task is provided, and only information generated by the engine is returned. Before submitting new context to the Probabilistic Engine, its session context is explicitly reset. The Probabilistic Engine strictly replies with its output to the Model Encapsulator that invoked it.

\begin{figure}
    \centering
    \includegraphics[width=1\linewidth]{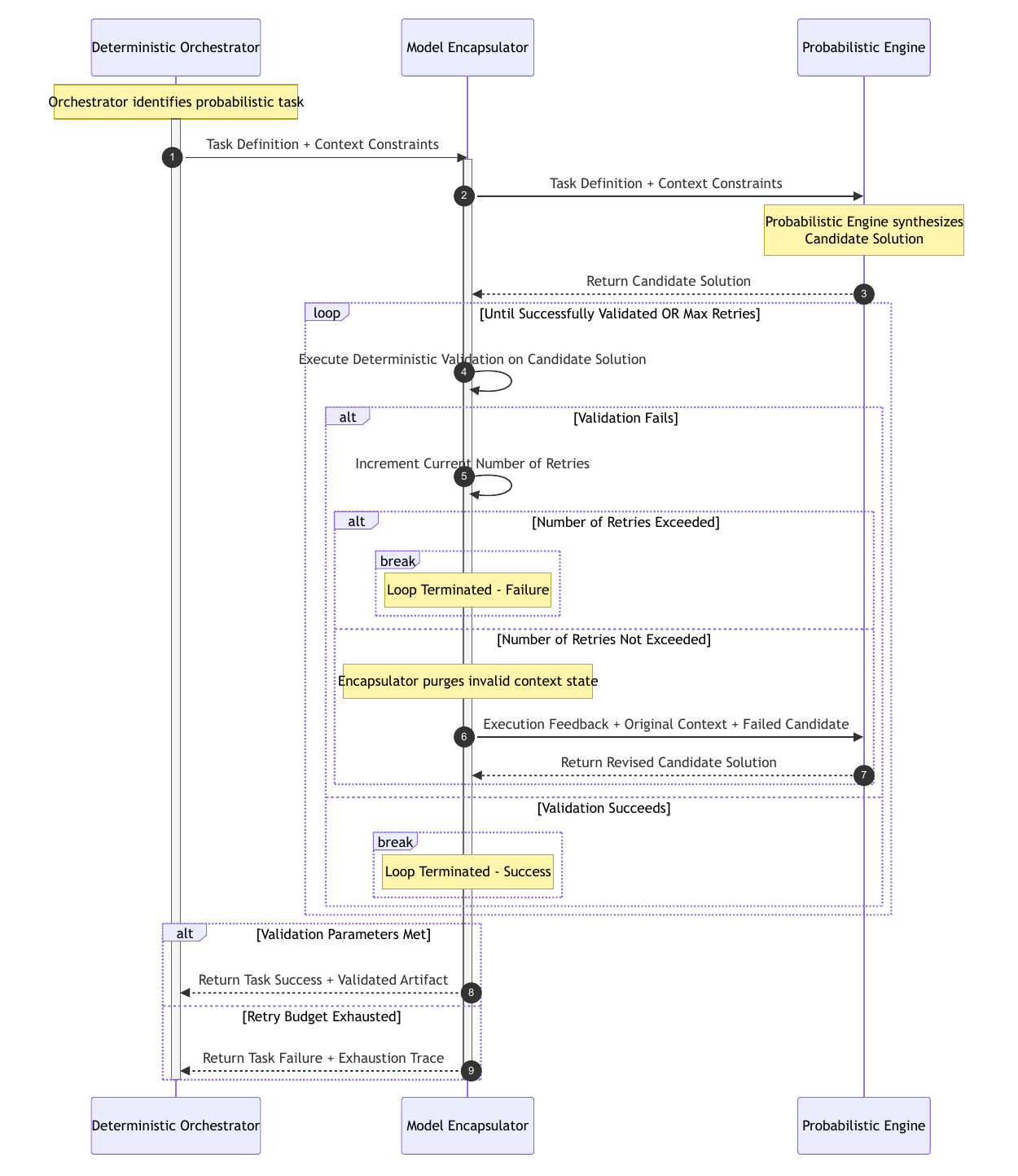}
    \caption{Validation Micro-Pattern Sequence Diagram}
    \label{fig:vmp}
\end{figure}

The Validation Micro-Pattern sequence diagram in Figure~\ref{fig:vmp} provides the definitive visual heuristic for Grounded Inference by exposing the strict boundary between state and probability. A primary heuristic for auditing generative architectures is the examination of sequence diagram lifelines. Continuous, "thick" activation is restricted entirely to the lifelines of instances of traditional computational primitives (i.e., Deterministic Orchestrator, Model Encapsulator, and State Registry). The Deterministic Orchestrator delegates the probabilistic task to the Model Encapsulator, which assumes full responsibility for managing the retry loop, tracking failure counts, and maintaining state. The Probabilistic Engine exists only as a series of disconnected flickers, spinning up to process a discrete payload, returning a candidate solution, and instantly terminating. If an architectural diagram requires a continuous activation lifeline on a Probabilistic Engine to function (unless explicitly representing an incremental streaming output), the architecture is state-leaking and fundamentally compromised.

\section{\textbf{Generative Model Risks}}

Generative models fundamentally resist traditional deterministic software testing and control. Even when treating the model as a static set of weights, multiple layers of operational, architectural, and mathematical instability inject non-reproducible variance into the system. Mitigating these risks requires structural guardrails built outside the model loop.

\subsection{Service Layer Instability and Hidden State Drift}

When utilizing a generative model via an external API or managed service layer, the execution environment operates as a black box, introducing hidden variables that destroy reproducibility. The pervasive assumption that setting a temperature to zero ensures determinism is demonstrably false in hosted environments. Empirical evaluations of API-based models configured for strict determinism reveal output accuracy variations of up to 15 percent across identical runs \cite{Atil2025}.

\begin{itemize}
    \item \textbf{Hidden System Prompt Injections:} Service providers may modify, wrap, or update system instructions (e.g., adding formatting constraints, safety guidelines, or few-shot examples) appended or pre-pended to the user prompt without the knowledge of the software engineer. Because self-attention dynamically calculates the relative significance of tokens at each generation step, even microscopic changes to hidden text (e.g., adding a trailing space or reordering system rules) alter the relative sequence positions of all subsequent tokens. This shifts their corresponding positional embeddings, fundamentally altering the dynamically computed attention distributions and yielding divergent outputs.
    \item \textbf{The Seed Illusion and Managed PRNG:} In standard computing, fixing a pseudorandom number generator (PRNG) seed guarantees an identical sequence of numbers across different executions. In hosted multi-tenant environments, providers dynamically optimize inference to maximize throughput via mechanisms like paged attention and dynamic batching. If the provider automatically resets, overrides, or interleaves the pseudo-random seed to manage parallel request loads, specifying a consistent seed via the API becomes ineffective, causing identical inputs to diverge across runs.
    \item \textbf{Unannounced Model and Alignment Updates:} Providers may ship minor patches or updated reinforcement learning (RLHF/RLAIF) alignments under frozen version aliases. These subtle modifications alter downstream token probabilities without changing the endpoint configuration.
\end{itemize}
\subsection{Distributed Architecture and Hardware-Level Non-Determinism}

Even if the service layer is entirely frozen (e.g., hosting a model locally on private infrastructure), low-level hardware constraints prevent bitwise determinism.

\begin{itemize}
    \item \textbf{Non-Associative Floating-Point Arithmetic and Distributed Compute:} Due to precision limitations and rounding errors, floating-point addition is non-associative: $(a + b) + c \neq a + (b + c)$. In massive GPU or TPU clusters, matrix multiplication is split across thousands of parallel threads. The exact order in which these distributed calculations arrive and are summed fluctuates dynamically based on network latency, cluster workload, and hardware thermal throttling. Yashwanth et.al. \cite{Yashwanth2025} demonstrated that this floating-point noise is not random, but highly structured and systemic, dictated by execution order and batch dimensions. 

Because generative models are auto-regressive, each generated token serves as the hard context for the next calculation. A microscopic variance in floating-point math at token 2 alters the log-probabilities of subsequent tokens. If a different token is selected early in the sequence, the error cascades exponentially, completely altering the trajectory of the output by token 50.
    \item \textbf{Backend Variability:} Research by Pape et al. \cite{Pape2026}  demonstrates that the choice of inference engine (such as vLLM, SGLang, llama.cpp, LMDeploy, and Ollama) acts as a hidden hyperparameter. Due to differences in hardware-specific parallel computing optimizations (e.g., custom CUDA kernels, prefix caching, and CUDA graphs) and engine-specific defaults in logit processing, the same model running on identical hardware can see benchmark scores swing by up to 16.6 percentage points solely due to backend architecture changes. The authors emphasize that this variance persists even under strictly standardized, greedy decoding settings (setting temperature to zero) across standard and reasoning-focused workloads, creating a consequential deployment gap between research validation and production reality.
\end{itemize}
\subsection{Intrinsic Stochastic Architecture}

At its core, the transformer architecture is a Probabilistic Engine rather than a discrete logic machine.

\begin{itemize}
    \item \textbf{Statistical Boundary Failures:} Generative models do not operate within a formal, bounded logical framework. They predict token sequences based on conditional probability distributions derived from training data. When encountering edge cases, rare logical syllogisms, or highly rigid formatting constraints, the model will inevitably cross statistical failure boundaries, generating hallucinations or structural deviations despite perfect inputs.
    \item \textbf{Sampling Distribution Realities:} Unless utilizing strict greedy decoding (setting temperature to zero), the model explicitly samples from a probability distribution. Any non-zero temperature setting guarantees that the software must handle a spectrum of semantic variants rather than a single predictable output.
\end{itemize}
Consequently, relying on zero-temperature generation or advanced prompting to enforce state boundaries is an architectural anti-pattern. If reproducibility could be achieved, it remains unsafe since even under strict deterministic constraints, the system remains subject to semantic volatility and logical hallucination. Securing generative architecture therefore requires external boundary enforcement. The Model Encapsulator ensures that the high-dimensional mapping of the Probabilistic Engine is structurally coerced into a rigid Interface Contract (such as a compiled Pydantic schema) before the Deterministic Orchestrator permits execution.

\section{\textbf{Discussion}}

While this manuscript critiques the autonomous application of Probabilistic Engines, conversational user interfaces represent a valid, distinct architectural pattern. However, this validity rests entirely on the presence of the human operator as a "Sovereign Validator". In a conversational interface, the human acts as the ultimate Model Encapsulator by absorbing the architectural risk.

Consider a probabilistic engine advising on the physical repair of 1,500 square feet of continuous hardwood flooring. The engine suggests that it is a simple task for a novice to spread liquid polyurethane uniformly across the entire surface simultaneously. If an autonomous execution plane received this instruction, it would blindly attempt the task. The human operator, however, possesses out-of-band physical context (e.g., the reality of airborne dust settling or curing times creating irreversible ridges) and asymmetric risk aversion. The operator inherently understands that attempting this without professional grounding could cause damage requiring remediation at a significantly higher financial cost than the original repair. Recognizing this logical failure, the human acts as the orchestrator and triggers a definitive halt condition.

Conversational AI is not an inherently safe probabilistic system. It is a highly volatile system stabilized strictly by the manual, physical grounding of a human operator functioning as an out-of-band Model Encapsulator. When transitioning to autonomous enterprise data pipelines where no human is present to intervene, the Probabilistic Engine must be subject to deterministic adherence to rigid constraints.

Tooling surrounding base Large Language Models is rapidly increasing in complexity. Commercial interfaces now employ multi-step inference, explode complex prompts into discrete sub-tasks, execute deterministic tools, and synthesize the results into a unified response. Furthermore, they actively integrate external search functionality for real-time grounding in current events. As these services mature, the deterministic boundaries codified in this framework will inevitably be pushed lower into the technology stack. Future commercial LLM offerings will likely present the illusion of a natively stateful, perfectly consistent engine. In reality, the foundational models will remain stateless and probabilistic. The platform providers will have simply integrated deterministic validation loops, epistemological grounding, and asynchronous state management directly into their managed orchestration layers. 

\section{\textbf{Conclusion}}

The architectural first-principles presented in this manuscript were chosen to lead to a single, overarching conclusion: Probabilistic Engines must be applied only to bridge gaps in the abilities of traditional computational methods. Although Probabilistic Engines are transforming the art-of-the-possible in the world of computing, their improper use can create foreseeable risks for enterprise data systems. Unfortunately, the Generative Anti-Patterns described herein are well represented in industry as of the writing of this manuscript. To control risks associated with incorporating generative models into traditional computing systems, this manuscript introduced a set of atomic primitives as well as specific design patterns to constrain the blast radius of behaviors including context collapse and stochastic variance.

\section{\textbf{Acknowledgments}}
The author utilized the Google Gemini 3 model family (including Gemini 3.1 Pro and Gemini 3.5 Flash) across three distinct phases of manuscript development. In the initial conceptual phase, Gemini was used via voice interfaces for unstructured exploratory dialogue and conceptual brainstorming. In the drafting phase, the model assisted in neutralizing spoken-text artifacts and formatting rough verbal dictations into structured prose. In the final compilation and editing phase, Gemini was utilized strictly as an external utility for targeted research verification, linguistic refinement, and logical consistency screening of the Grounded Inference primitives. The author reviewed, verified, and accepts full, unshared accountability for all final architecture, technical assertions, and textual content.

\bibliographystyle{unsrt}  
\bibliography{references} 

\end{document}